\documentclass{article}

\usepackage{microtype}
\usepackage{graphicx}
\usepackage{subcaption}
\usepackage{booktabs} 

\usepackage{hyperref}

\usepackage[preprint]{icml2026}

\usepackage{amsmath}
\usepackage{amssymb}
\usepackage{mathtools}
\usepackage{amsthm}
\usepackage[dvipsnames]{xcolor}
\definecolor{DeepGreen}{RGB}{0,120,0} 
\definecolor{DeepRed}{RGB}{150,0,0}   
\usepackage{float}  

\usepackage{graphicx}
\usepackage{multirow} 
\usepackage{listings}
\usepackage[capitalize,noabbrev,nameinlink]{cleveref}

\theoremstyle{plain}

\theoremstyle{definition}

\theoremstyle{remark}

\usepackage[textsize=tiny]{todonotes}

\icmltitlerunning{When Domains Interact: Asymmetric and Order-Sensitive Cross-Domain Effects in Reinforcement Learning for Reasoning}

\begin{document}

\twocolumn[
  \icmltitle{When Domains Interact: Asymmetric and Order-Sensitive Cross-Domain Effects in Reinforcement Learning for Reasoning}



  \icmlsetsymbol{equal}{*}

  \begin{icmlauthorlist}
    \icmlauthor{Wang  Yang}{case}
    \icmlauthor{Shouren Wang}{case}
    \icmlauthor{Chaoda Song}{case}
    \icmlauthor{Chuang Ma}{kyoto}
    \icmlauthor{Xinpeng Li}{case}
    \icmlauthor{Nengbo Wang}{case}\\
    \icmlauthor{Kaixiong Zhou}{nc}
    \icmlauthor{Vipin Chaudhary}{case}
    \icmlauthor{Xiaotian Han}{case}
  \end{icmlauthorlist}

  \icmlaffiliation{case}{Case Western Reserve University}
  \icmlaffiliation{nc}{North Carolina State University}
  \icmlaffiliation{kyoto}{NII, Japan}

  \icmlcorrespondingauthor{Vipin Chaudhary}{vxc204@case.edu}
  \icmlcorrespondingauthor{Xiaotian Han}{xxh584@case.edu}

  \icmlkeywords{Machine Learning, ICML}

  \vskip 0.3in
]



\printAffiliationsAndNotice{}  

\begin{abstract}

Group Relative Policy Optimization (GRPO) has become a key technique for improving reasoning abilities in large language models, yet its behavior under different domain sequencing strategies is poorly understood. In particular, the impact of sequential (one domain at a time) versus mixed-domain (multiple domain at a time) training in GRPO has not been systematically studied. We provide the first systematic analysis of training-order effects across math, science, logic, and puzzle reasoning tasks. We found (1) single-domain generalization is highly asymmetric: training on other domains improves math reasoning by approximately 25\% accuracy, while yielding negligible transfer to logic and puzzle; (2) cross-domain interactions are highly order-dependent: training in the order math$\rightarrow$science achieves 83\% / 41\% accuracy on math / science, while reversing the order to science$\rightarrow$math degrades performance to 77\% / 25\%; (3) no single strategy is universally optimal in multi-domain training: sequential training favors math (up to 84\%), mixed training favors science and logic,
and poor ordering can incur large performance gaps (from 70\% to 56\%). Overall, our findings demonstrate that GRPO under multi-domain settings exhibits pronounced asymmetry, order sensitivity, and strategy dependence, highlighting the necessity of domain-aware and order-aware training design. Our code is available at \url{https://github.com/uservan/cross_domain}.

\end{abstract}

\section{Introduction}

\begin{figure}[ht]
  \centering
  \vspace{-0.1in}

  \begin{subfigure}[t]{0.48\linewidth}
    \centering
    \includegraphics[width=\linewidth]{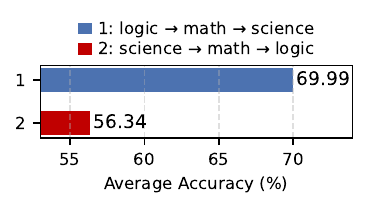}\vspace{-5pt}
    \caption{Training-order comparison across multi-domains.}
    \label{fig:comparison}
  \end{subfigure}
    \hfill
  \begin{subfigure}[t]{0.48\linewidth}
    \centering
    \includegraphics[width=\linewidth]{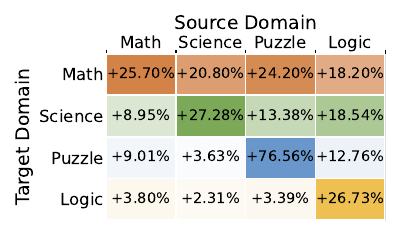}\vspace{-5pt}
    \caption{Single-domain generalization transfer.}
    \label{fig:transfer_graph}
  \end{subfigure}

  \vspace{0.6em}

  \begin{subfigure}[t]{0.48\linewidth}
    \centering
    \includegraphics[width=\linewidth]{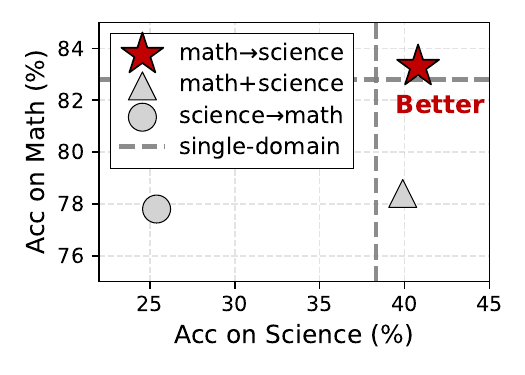}\vspace{-5pt}
    \caption{Math $\leftrightarrow$ Science.}
    \label{fig:math_science_transfer}
  \end{subfigure}
  \hfill
  \begin{subfigure}[t]{0.48\linewidth}
    \centering
    \includegraphics[width=\linewidth]{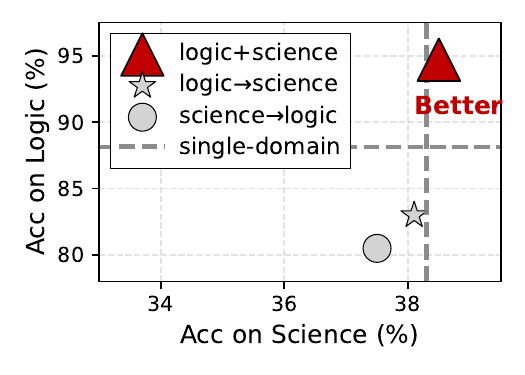}\vspace{-5pt}
    \caption{Logic $\leftrightarrow$ Science.}
    \label{fig:logic_science_transfer}
  \end{subfigure}

  \vspace{-0.3em}
  \caption{
  \textbf{Cross-domain transfer and interference in GRPO.} \textbf{(a)} Training-order comparison across domains. Opposite sequential training orders lead to large differences in average multi-domain performance. \textbf{(b)} Single-domain transfer graph. Values (e.g., $+20.8\%$) denote accuracy gains when training on a source domain (science) and evaluating on a target domain (math); darker colors indicate larger gains. Math shows strong cross-domain transfer, science moderate transfer, and logic/puzzle minimal transfer. \textbf{(c) and (d)} Asymmetric two-domain transfer. "$\rightarrow$" means sequential training; “+” denotes mixed training; dashed lines indicate single-domain baselines; points closer to the top-right achieve better joint performance across both domains. Training order is critical: math$\rightarrow$science yields substantially higher accuracy, while science interferes with logic in sequential training.
  }
  \label{fig:overview_transfer}
  \vspace{-6pt}
\end{figure}

\begin{figure*}[t]
  \centering
  \vspace{-0.1in}

  \begin{subfigure}[t]{0.3\linewidth}
    \centering
    \includegraphics[width=\linewidth]{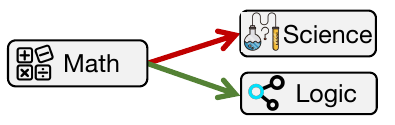}
    \caption{Single-domain generalization}
    \label{fig:overview_single}
  \end{subfigure}
  \hfill
  \begin{subfigure}[t]{0.38\linewidth}
    \centering
    \includegraphics[width=\linewidth]{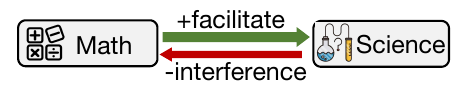}
    \caption{Cross-domain interaction}
    \label{fig:overview_two}
  \end{subfigure}
  \hfill
  \begin{subfigure}[t]{0.3\linewidth}
    \centering
    \includegraphics[width=\linewidth]{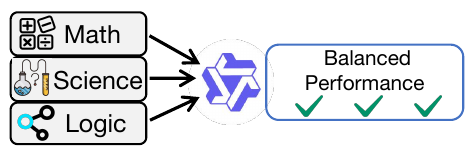}
    \caption{Multi-domain training}
    \label{fig:overview_multi}
  \end{subfigure}

 \vspace{-6pt}
  \caption{Overview of our systematic analysis of GRPO in multi-domain settings.
  \textbf{(a)} Single-domain generalization: how training on one domain transfers to others.
  \textbf{(b)} Cross-domain interaction: asymmetric facilitation and interference between two domains under different training orders.
  \textbf{(c)} Multi-domain training: different training strategies for achieving improved and balanced performance across domains.
  }
  \label{fig:grpo_multidomain_overview}
\end{figure*}


Most existing improvements to Group Relative Policy Optimization (GRPO)~\cite{shao2024deepseekmath, liu2024deepseek} have been primarily developed and evaluated in the math and code domains, such as DAPO~\cite{yu2025dapo}, Dr.\ GRPO~\cite{liu2025understanding}, GSPO~\cite{zheng2025group}, etc. Existing applications of GRPO to other domains generally study each domain independently, such as logic~\cite{xie2025logicrlunleashingllmreasoning} and puzzles~\cite{chen2025enigmata}. 

In contrast, the behavior of GRPO under multi-domain training has received little attention. It remains unclear how domain-specific signals interact during multi-domain training. As shown in \Cref{fig:comparison}, opposite training orders across domains surprisingly can yield evaluation results ranging from high to low accuracy, indicating that multi-domain GRPO may be highly sensitive to domain interactions. This motivates a systematic study of cross-domain effects in reinforcement learning for reasoning training.



In \Cref{fig:grpo_multidomain_overview}, we first systematically explore and analyze the behavior of Group Relative Policy Optimization in multi-domain settings. Our study is organized around three main aspects: (1) \textit{single-domain generalization}: we examine how single-domain training affects reasoning performance in
other domains. (2) \textit{cross-domain interaction}: we investigate cross-domain interactions between two domains, asking whether reasoning ability in a given domain is enhanced or degraded when training on another domain. (3) \textit{multi-domain training}: we study how to directly train on multiple domains to achieve improved and balanced performance across all domains. To facilitate this analysis, we construct training and evaluation datasets covering four representative domains: mathematics, science, logic, and puzzle.
\begin{table}[t]
\vspace{-0.1in}
\caption{
Accuracy comparison of representative training strategies of multi-domain training. Sequential strategies exhibit large performance variations—for example, logic$\rightarrow$math$\rightarrow$science substantially outperforms science$\rightarrow$math$\rightarrow$logic.
Moreover, different domains favor different training paradigms: math benefits
most from the optimal sequential strategy, while science and other domains
achieve better performance under mixed training.
}
\label{tab:intro_seq_vs_mixed_split}
\vspace{-6pt}
\centering
\small
\setlength{\tabcolsep}{8pt}
\renewcommand{\arraystretch}{1.15}
\begin{tabular}{c|ccc}
\toprule







Training Strategy
& Math & Science & Logic \\
\midrule
single-domain
& 82.76 & 38.28 & 88.14 \\
\midrule
logic $\rightarrow$ math $\rightarrow$ science
& \textbf{84.00} & 36.67 & 87.31 \\

science $\rightarrow$ math $\rightarrow$ logic
& 78.24 & 26.97 & 63.80 \\

math + science + logic
& 80.44 & \textbf{40.00} & \textbf{95.37} \\

\bottomrule
\end{tabular}
\end{table}

The results of single-domain generalization are shown in \Cref{fig:transfer_graph}. Models are trained on one domain (train field) and evaluated on another (test field). A larger accuracy improvement (final-step accuracy minus first-step accuracy) on the test field is indicated by a darker color. We observe that: (1) mathematical reasoning ability can be effectively activated by training on data from almost all other domains; (2) science reasoning can also benefit from cross-domain training, though to a lesser extent than math; and (3) logic and puzzle reasoning abilities are largely difficult to acquire from other domains and show minimal cross-domain gains.

The cross-domain results are summarized in \Cref{fig:math_science_transfer} and \Cref{fig:logic_science_transfer}.
We observe that: (1) \textit{training order plays a critical role for the math and science domains}. Training in the order math$\rightarrow$science achieves the best performance on both math and science, whereas the reverse order science$\rightarrow$math leads to substantial degradation in both domains. Mixed training improves overall performance to some extent, but remains inferior to the math$\rightarrow$science ordering. (2) \textit{Science training consistently interferes with logical reasoning}. Under both science$\rightarrow$logic and logic$\rightarrow$science, model performance on the logic domain decreases. Notably, mixed training on logic and science effectively mitigates this interference and yields improved accuracy on both domains.


The results of multi-domain training are reported in \Cref{tab:intro_seq_vs_mixed_split}. We compare single-domain training, the best and worst sequential training strategies in terms of average performance, and mixed training. We observe that: (1) \textit{in sequential training, domain order has a substantial impact on performance}. For example, the sequence science$\rightarrow$math$\rightarrow$logic achieves accuracies of 26.67 and 63.8 on the science and logic domains, respectively, which are far below those obtained by the reverse order logic$\rightarrow$math$\rightarrow$science. (2) \textit{Different domains favor different training strategies}. For the math domain, the optimal sequential strategy yields the highest accuracy,
reaching up to 84\%. In contrast, for science and logic, mixed training is more suitable and leads to the best overall performance on these domains.



Overall, our contributions can be summarized as follows:
\begin{itemize}
  \item We show that domain interactions in GRPO are highly asymmetric and strongly dependent on training order.
  \item We provide practical guidance for multi-domain GRPO by identifying when training order is critical and when mixed training is preferable. Ignoring these effects can introduce substantial bias, with worst-case sequential strategies performing far worse than optimal ones (e.g., 70\% vs.\ 56\% average accuracy).
  \item We present the first systematic investigation of GRPO under multi-domain training, laying a foundation for future work on principled multi-domain reinforcement learning for reasoning models.
\end{itemize}

\section{Single-Domain Generalization}


This section first examines the cross-domain impact of domain-specific training. Specifically, we study how training on data from one domain (like math) affects performance in other domains (like science). Our experimental results show that the math and science domains are more susceptible to interference from other domains, whereas puzzle and logic
domains are comparatively robust.


\textbf{Experimental Setup}. We adopt the standard Group Relative Policy Optimization~\cite{shao2024deepseekmath,liu2024deepseek}.
The batch size is set to 256, the learning rate to $1\times10^{-6}$, and the model
is trained for 15 epochs.
For each problem, we generate 8 candidate responses, with the maximum output
length capped at 16K tokens.
Due to computational resource constraints, all experiments are conducted using
the Qwen3-4B-Base model~\cite{yang2025qwen3}.

\begin{table}[t]
\vspace{-0.1in}
\centering
\caption{Overview of Training Datasets Across Four Domains. "test" means use the test data from the Dataset Source.}
\vspace{-6pt}
\label{tab:dataset_overview}
\setlength{\tabcolsep}{5pt}
\small
\begin{tabular}{cccc}
\toprule
\textbf{Domain} & \textbf{Dataset Source} & \textbf{Size} & \textbf{Test Set} \\
\midrule
Math 
& Skywork-OR1-RL-Data
& 5K 
& MATH500 \\

Science 
& OpenScienceReasoning-2 
& 5K 
& GPQA \\

Logic 
& knights-and-knaves dataset 
& 5K 
& test \\

Puzzle 
& Enigmata-Data 
& 5K 
& test \\
\bottomrule
\end{tabular}
\end{table}
\subsection{Dataset Construction}


We first introduce the four domains considered in this work, namely
mathematics, science, logic, and puzzle.
\Cref{tab:dataset_overview} summarizes the data sources and the corresponding
data sizes for each domain.

\textbf{Mathematics.} The mathematics data are drawn from the Skywork/Skywork-OR1-RL-Data dataset~\cite{he2025skywork, skywork-or1-2025},
  which consists of verifiable, challenging, and diverse mathematics problems (105K) as well as coding questions (14K). We select 5K mathematics problems based on the difficulty levels annotated using DeepSeek-R1-Distill-Qwen-7B, ensuring that the selected problems are well matched to the capacity of Qwen3-4B-Base.

  \textbf{Science}. Nvidia/OpenScienceReasoning-2~\cite{NemotronPostTrainingDatasetV1,bercovich2025llamanemotronefficientreasoningmodels} is the data source of science, which contains both multiple-choice and open-ended question--answer pairs with
  detailed reasoning traces across a wide range of domains, including STEM, law, economics, and humanities.
  In this work, we retain only multiple-choice questions from the STEM domain, each with a single correct answer, to ensure consistency and uniformity in supervision.

  \textbf{Logic.} The dataset is taken from the K-and-K/knights-and-knaves dataset~\cite{xie2024memorization}, which is a benchmark for evaluating logical reasoning capabilities of
  large language models.
  We sample 5K instances in total, and use 700 examples from the original test
  split as the evaluation set.

\textbf{Puzzle.} For the puzzle domain, we use data generated by the Enigmata-Data project~\cite{chen2025enigmata},
  which has 36 distinct task types spanning 7 categories of logical reasoning puzzles. We construct training sets totaling 5K examples, covering easy, medium, and hard difficulty levels.





We evaluate performance in the mathematics and science domains using
MATH500~\cite{lightman2023lets} and GPQA~\cite{rein2024gpqa}, respectively.
For the logic and puzzle domains, evaluation is conducted using subsets selected
from their original test sets.

\begin{table}[t]
\vspace{-0.1in}
\caption{
    Performance comparison on MATH500.
    We report accuracy at step~1 (Begin), accuracy at the final step (Last),
    the absolute improvement ($\Delta$, in percentage points), and the accuracy rank.
}
\label{tab:math}
\vspace{-6pt}
\centering
\setlength{\tabcolsep}{6pt}
\begin{tabular}{c|cccc}
\toprule
Train Field
& Begin (\%)
& Last (\%)
& $\Delta$
& Rank \\
\midrule
Math    & 40.6 & 66.3 & +25.7 & 1 \\
Science & 41.8 & 62.6 & +20.8 & 2 \\
Logic   & 36.4 & 60.6 & +24.2 & 3 \\
Puzzle  & 40.2 & 58.4 & +18.2 & 4 \\
\bottomrule
\end{tabular}
\end{table}
\begin{figure}[t]
  \centering
  \vspace{-0.1in}
  \includegraphics[width=0.8\linewidth]{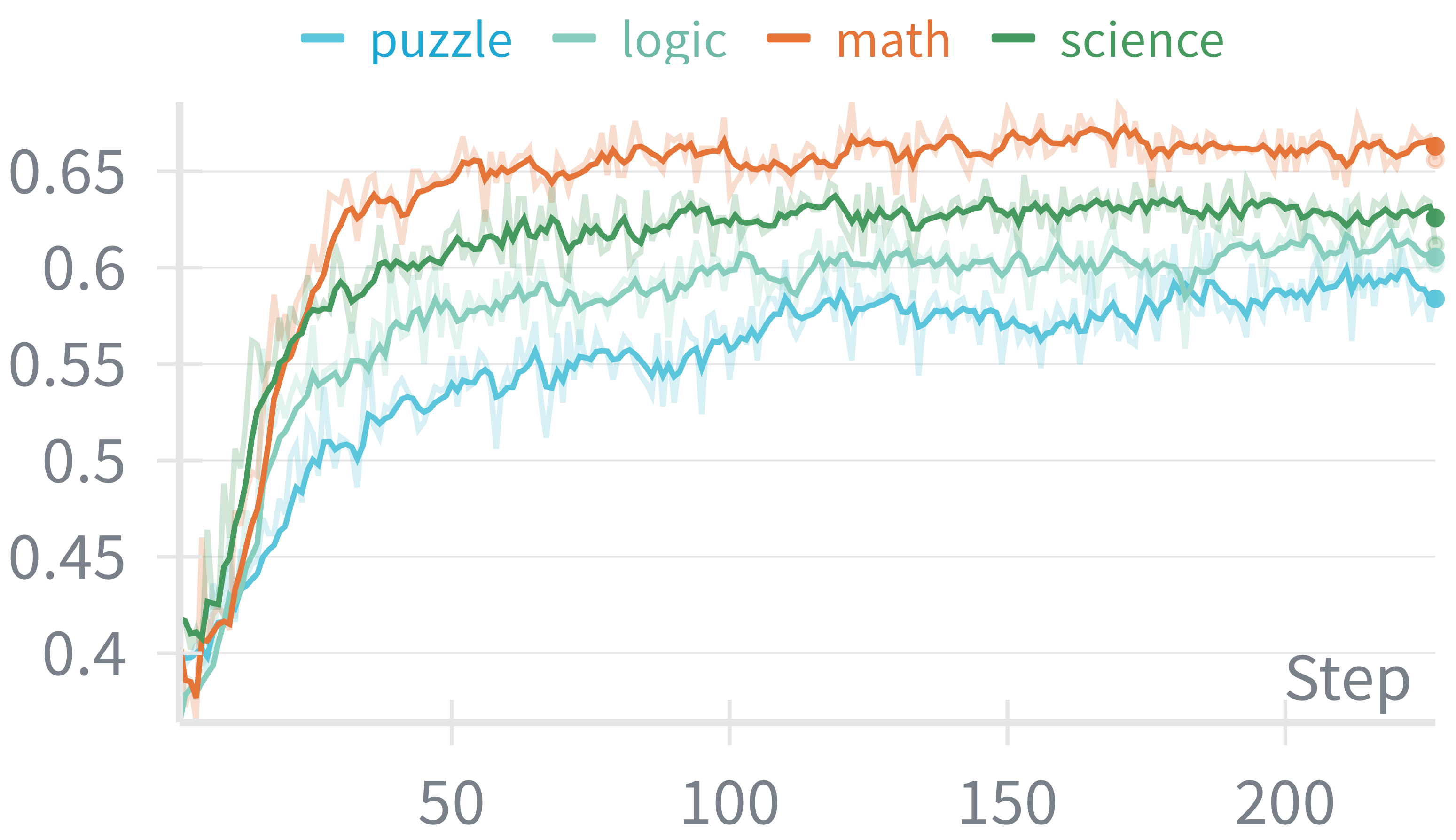}
  \vspace{-6pt}
  \caption{
    Accuracy on MATH500 at each training step for Qwen3-4B-Base when trained with
    GRPO on different domain data (math, logic, puzzle and science). Across all domains, training rapidly activates the model’s mathematical
    reasoning ability.
    }
  \label{fig:math}
\end{figure}

\subsection{Mathematical Reasoning is Highly Transferable Across Domains}

\Cref{fig:math} illustrates the accuracy on MATH500 across training steps for models trained on data from four different domains, as recorded in Weights \& Biases. The accompanying \Cref{tab:math} summarizes the accuracy at the beginning (step~1) and at the final training step, as well as the final ranking of the four training domains on MATH500.


Regardless of the training domain, the Qwen3-4B-Base model exhibits a consistent improvement on MATH500 after training, with an absolute accuracy gain of approximately 20 percentage points. This indicates that data from all four domains can effectively enhance the
model’s mathematical reasoning ability. Among them, training on puzzle data yields the smallest improvement, whereas
training on mathematics data achieves the best final performance.

\begin{table*}
\caption{
    Performance comparison across GPQA, Logic Test, and Puzzle Test on different training datasets. For each train field, we report accuracy at step~1 (Begin), accuracy at the final step (Last), the absolute improvement ($\Delta$,
    in percentage points), and the accuracy rank.
}
\label{tab:science，puuzle and logic}
\vspace{-6pt}
\centering
\setlength{\tabcolsep}{6pt}
\begin{tabular}{c|cccc|cccc|cccc}
\toprule
\multirow{2}{*}{Train Field}
& \multicolumn{4}{c|}{GPQA}
& \multicolumn{4}{c|}{Puzzle Test}
& \multicolumn{4}{c}{Logic Test} \\
\cmidrule(lr){2-5} \cmidrule(lr){6-9} \cmidrule(lr){10-13}
& Begin & Last & $\Delta$ & Rank
& Begin & Last & $\Delta$ & Rank
& Begin & Last & $\Delta$ & Rank \\
\midrule
Math
& 10.66 & 19.61 & +8.95 & 4
& 2.05 & 5.85 & +3.80 & 3
& 21.29 & 30.30 & +9.01 & 3 \\

Science
& 9.64 & 36.92 & +27.28 & 1
& 1.54 & 3.85 & +2.31 & 4
& 22.29 & 25.92 & +3.63 & 4 \\

Logic
& 8.12 & 21.50 & +13.38 & 3
& 2.82 & 6.21 & +3.39 & 2
& 21.14 & 97.70 & +76.56 & 1 \\

Puzzle
& 10.66 & 29.20 & +18.54 & 2
& 2.31 & 29.04 & +26.73 & 1
& 20.14 & 32.90 & +12.76 & 2 \\
\bottomrule
\end{tabular}
\end{table*}
\begin{figure*}
  \centering
  \begin{minipage}[t]{0.32\textwidth}
    \centering
    \includegraphics[width=\linewidth]{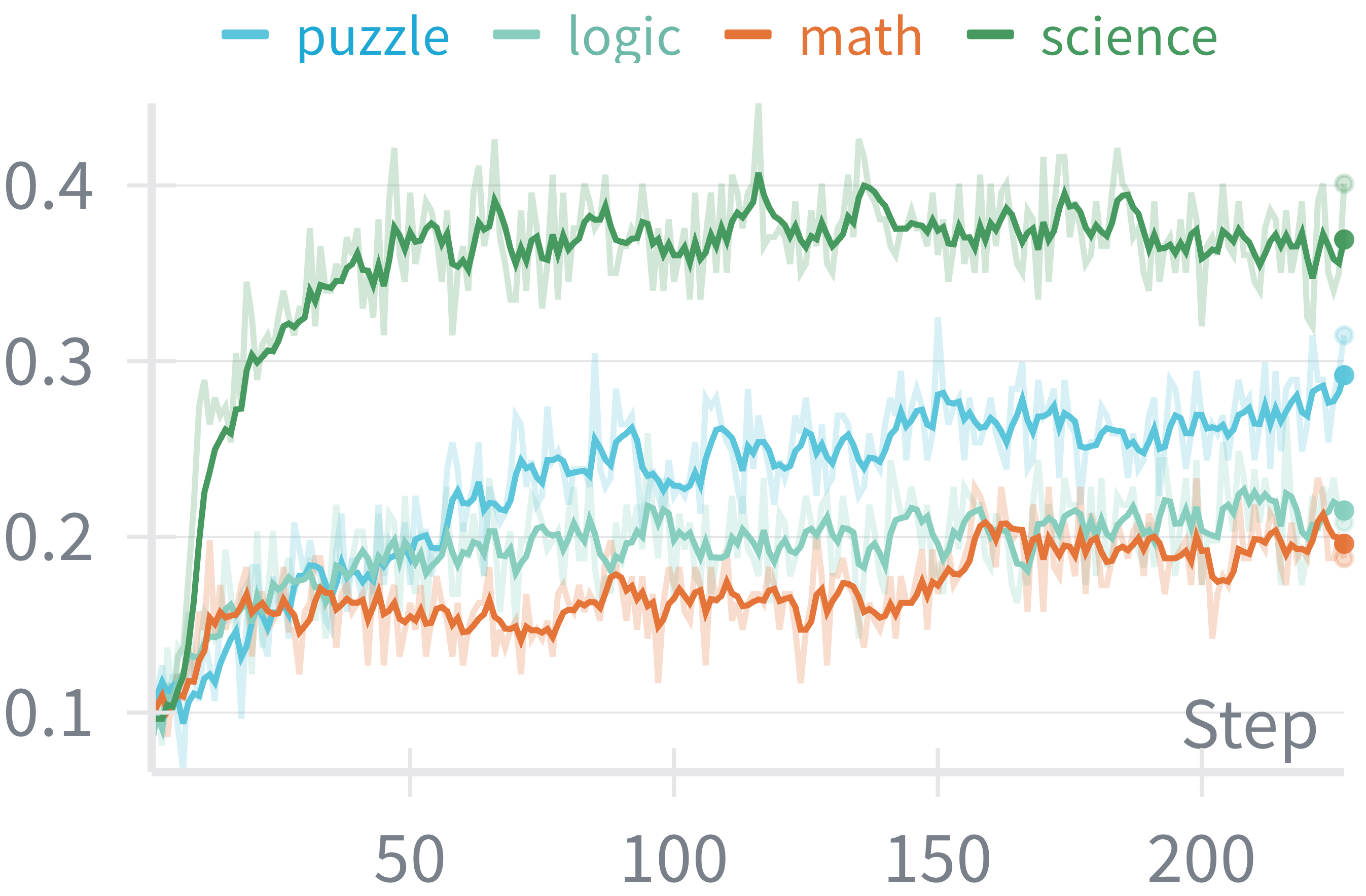}
    \\[-0.3em]
    {(a) science}
  \end{minipage}
  \hfill
  \begin{minipage}[t]{0.32\textwidth}
    \centering
    \includegraphics[width=\linewidth]{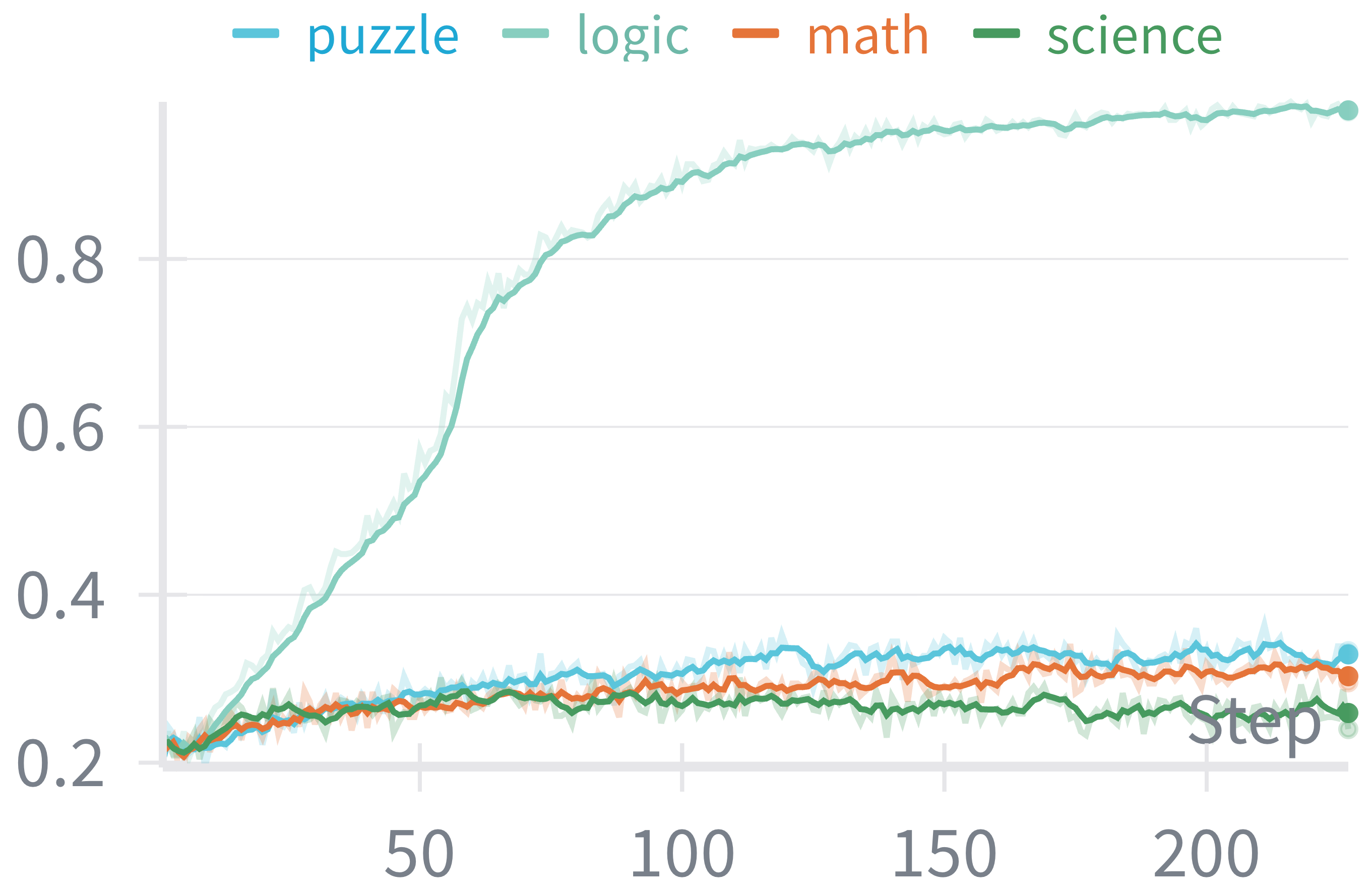}
    \\[-0.3em]
    { (b) logic}
  \end{minipage}
  \hfill
  \begin{minipage}[t]{0.32\textwidth}
    \centering
    \includegraphics[width=\linewidth]{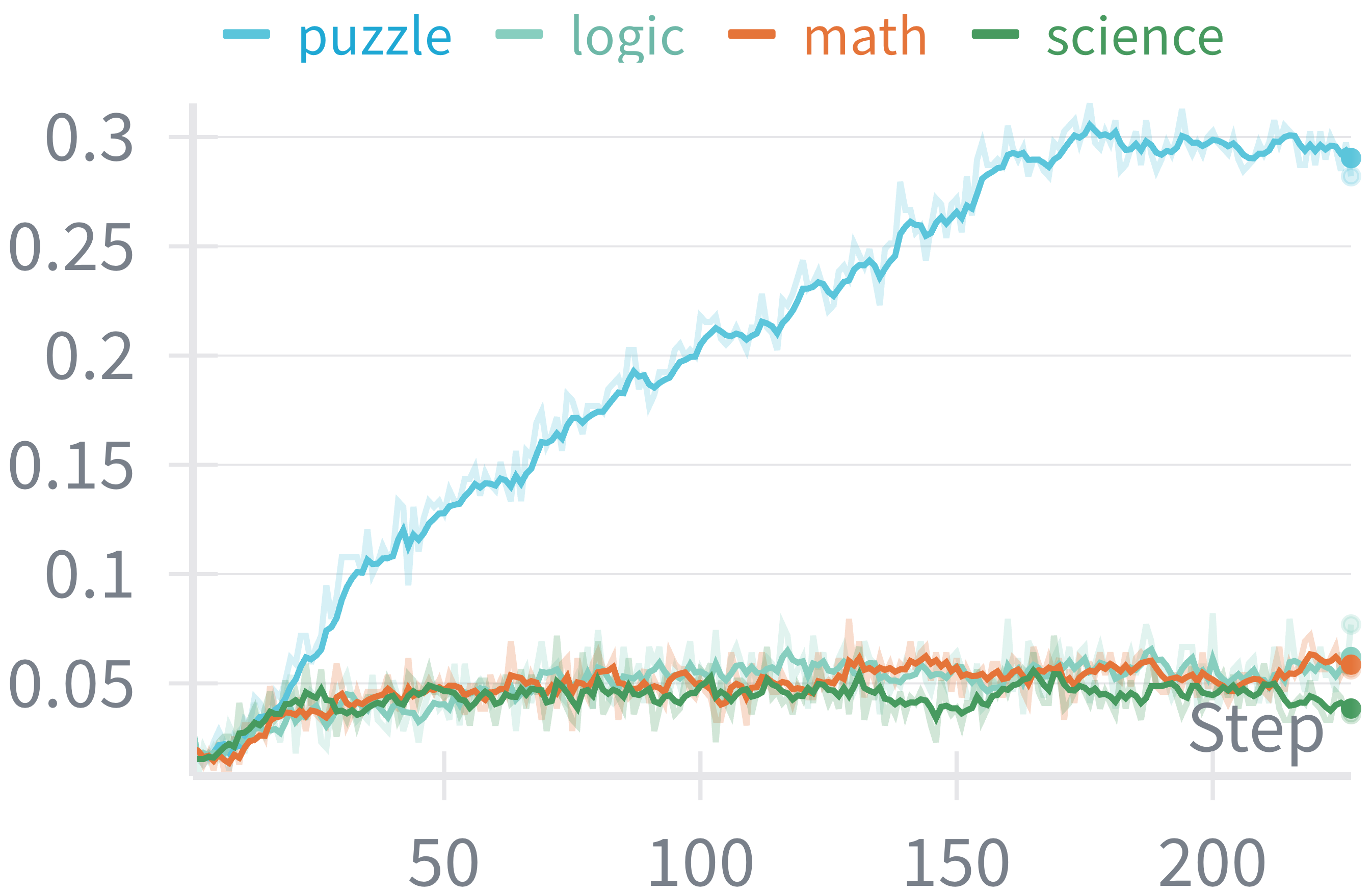}
    \\[-0.3em]
    {(c) puzzle}
  \end{minipage}
  \vspace{-3pt}
  \caption{
    Step-wise accuracy of Qwen3-4B-Base on GPQA, Logic Test, and Puzzle Test during GRPO training with data from different domains (a) Results on the science domain (GPQA) show that science reasoning can be partially activated by training on other domains, though less readily than in math. (b) and (c) show results on the logic and puzzle domains respectively. The model’s logic and puzzle reasoning abilities are largely insensitive to training on other domains and are primarily activated by in-domain data..
    }
  \label{fig:others}
\end{figure*}


We hypothesize that this behavior stems from the exposure of Qwen3-4B-Base to substantial mathematical data during the pretraining stage.
As a result, during the subsequent GRPO-based reinforcement learning phase, training on data from different domains can still transfer beneficially to the
model’s mathematical capability.

\begin{figure}[t]
  \centering
  \vspace{-0.08in}
  \includegraphics[width=0.95\linewidth]{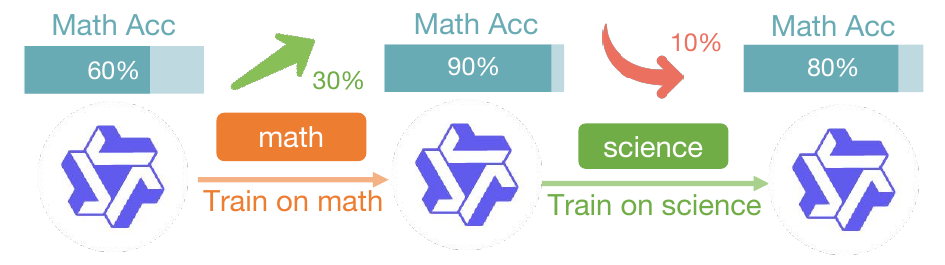}
  \vspace{-6pt}
  \caption{
    Training pipeline for Cross-Domain interaction. We first train the model on Field~1 (math) to obtain $M_1$, then continue training on Field~2 (science) to obtain $M_2$, and finally evaluate Field~1 performance to quantify forgetting/transfer: $\Delta_{\mathrm{F1}\leftarrow \mathrm{F2}}=\mathrm{Acc}_{\mathrm{F1}}(M_2)-\mathrm{Acc}_{\mathrm{F1}}(M_1)$.
  }
  \label{fig:seq_train}
\end{figure}

\subsection{Scientific Reasoning is Moderately Transferable and Domain-Sensitive}


Similar to the results on MATH500, Qwen3-4B-Base model exhibits performance improvements on GPQA when trained on data from all four domains, indicating that
cross-domain training can also enhance scientific reasoning ability. However, the magnitude of improvement on GPQA is less consistent and generally
smaller than that observed on MATH500.


\begin{table*}[t]
\caption{
Two-domain training results across four domains: math, science, logic, and puzzle.
For each target domain, \emph{original} denotes the performance of a model trained
only on the target domain.
\emph{Continue Training Field} reports performance after the model is further
trained on another domain following target-domain training
(\textit{target}$\rightarrow$\textit{other}),
while \emph{Before Training Field} reports performance when the model is first
trained on another domain and then trained on the target domain
(\textit{other}$\rightarrow$\textit{target}).
We report accuracy (\%) and the performance change
$\Delta=\text{Acc}_{\text{current}}-\text{Acc}_{\text{original}}$
(in percentage points).
}
\label{tab:four_domain_one_row}
\vspace{-6pt}
\centering
\setlength{\tabcolsep}{6pt}
\begin{tabular}{c|c|c|ccc|ccc}
\toprule
Target Domain & &
& \multicolumn{3}{c|}{Domain $\rightarrow$ Continue Training Field}
& \multicolumn{3}{c}{Before Training Field $\rightarrow$ Domain} \\
\midrule

\multirow{3}{*}{Math}
& Metric & original
& science & logic & puzzle
& science & logic & puzzle \\
\cmidrule(lr){2-3} \cmidrule(lr){4-6} \cmidrule(lr){7-9}
& Accuracy (\%) & 82.8
& 83.3 & 82.2 & 78.8
& 77.8 & 82.9 & 79.7 \\
& Delta (\%) & --
& +0.5 & -0.6 & \textcolor{DeepRed}{-4.0}
& \textcolor{DeepRed}{-5.0} & +0.1 & \textcolor{DeepRed}{-3.1} \\

\midrule

\multirow{3}{*}{Science}
& Metric & original
& mathematics & logic & puzzle
& mathematics & logic & puzzle \\
\cmidrule(lr){2-3} \cmidrule(lr){4-6} \cmidrule(lr){7-9}
& Accuracy (\%) & 38.3
& 25.4 & 37.5 & 37.5
& 40.8 & 38.1 & 39.0 \\
& Delta (\%) & --
& \textcolor{DeepRed}{-12.9} & -0.8 & -0.8
& \textcolor{DeepGreen}{+2.5} & -0.2 & +0.7 \\

\midrule

\multirow{3}{*}{Logic}
& Metric & original
& mathematics & science & puzzle
& mathematics & science & puzzle \\
\cmidrule(lr){2-3} \cmidrule(lr){4-6} \cmidrule(lr){7-9}
& Accuracy (\%) & 88.1
& 86.0 & 83.0 & 88.1
& 87.8 & 80.5 & 87.0 \\
& Delta (\%) & --
& -2.1& \textcolor{DeepRed}{-5.1} & 0.0
& -0.3 & \textcolor{DeepRed}{-7.6} & -1.1 \\

\midrule

\multirow{3}{*}{Puzzle}
& Metric & original
& mathematics & science & logic
& mathematics & science & logic \\
\cmidrule(lr){2-3} \cmidrule(lr){4-6} \cmidrule(lr){7-9}
& Accuracy (\%) & 20.2
& 19.7 & 19.1 & 20.5
& 18.5 & 21.9 & 20.3 \\
& Delta (\%) & --
& -0.5 & -1.1 & +0.3
& -1.7 & +1.7 & +0.1 \\

\bottomrule
\end{tabular}
\end{table*}
\begin{figure*}[t]
  \centering

  \begin{subfigure}{0.49\textwidth}
    \centering
    \includegraphics[width=\linewidth]{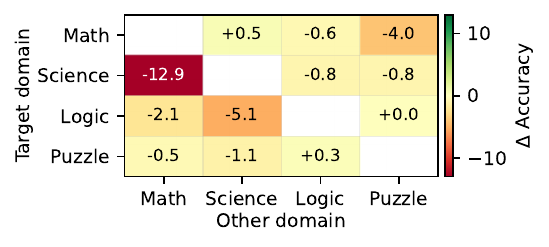}\\[-8pt]
    \caption{Training Order: target→other}
    \label{fig:heatmap_continue}
  \end{subfigure}
    \hfill
  \begin{subfigure}{0.49\textwidth}
    \centering
    \includegraphics[width=\linewidth]{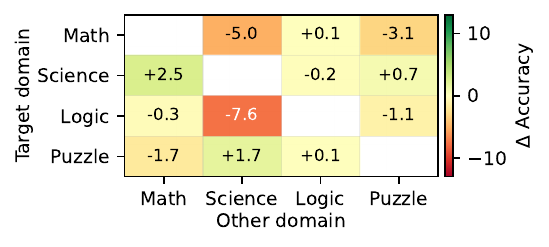}\\[-8pt]
    \caption{Training Order: other→target}
    \label{fig:heatmap_before}
  \end{subfigure}
    \vspace{-6pt}
  \caption{
    Asymmetric cross-domain transfer and interference effects.Each cell reports $\Delta=\mathrm{Acc}_{\mathrm{current}}-\mathrm{Acc}_{\mathrm{original}}$ (in percentage points). For example, in (a), $-12.9$ indicates the performance drop on the science domain when the model is first trained on science and then further trained on math, compared to training on science alone. In (b), $+2.5$ indicates the performance gain on the science domain when the model is first trained on math and then trained on science, relative to the science-only baseline.
    }
  \label{fig:heatmap_transfer}
\end{figure*}

As shown in \Cref{tab:science，puuzle and logic} and \Cref{fig:others}, the gains in scientific performance vary substantially across training domains. For example, training on science data yields an accuracy improvement of approximately 27 percentage points on GPQA, whereas training on mathematics data
results in a much smaller gain of around 9 percentage points.
This contrast highlights the higher sensitivity of scientific reasoning
performance to the choice of training data.


We conjecture that the model’s pretraining corpus already includes a certain amount of science-related data, which enables GRPO training on different domains to transfer to scientific reasoning to some extent. Nevertheless, the relatively smaller and less uniform gains suggest that the quantity and coverage of science data during pretraining are likely more limited
than those of mathematics data.

\subsection{Puzzle and Logic are Largely Domain-Specific}

In contrast to the observations on the mathematics and science domains, the results on the puzzle and logic domains exhibit a markedly different pattern, with substantially weaker cross-domain transfer effects on logic and puzzle test, as shown in both \Cref{tab:science，puuzle and logic} and \Cref{fig:others}.


On the logic test, only training on logic-domain data enables the model to reach a high accuracy of approximately 97\%; training on puzzle data yields a moderate improvement of around 30\%, while training on mathematics or science data results in much smaller gains, with other domains largely plateauing at roughly 30\% accuracy at the final step.


A similar trend is observed on the puzzle test: except for training on puzzle-domain data, which achieves an accuracy of around 30\%, training on other domains brings little to no improvement on the puzzle test.


We hypothesize that these domains are underrepresented during the model’s pretraining stage, requiring domain-specific reinforcement learning to effectively improve performance.




\section{Cross-Domain Interaction}


The phenomenon of the previous section raises a natural question: does training on data from Domain~B, either before or after a model acquires reasoning ability in Domain~A, affect learning and retention of Domain~A reasoning ability? 




\textbf{Experimental Setup}. To investigate domain-wise forgetting or retention before and after training on
other domains, we design the pipeline illustrated in \Cref{fig:seq_train}. Specifically, the model is first trained on Domain~1 (e.g., mathematics) to obtain Model~1, and is then further trained on Domain~2 to obtain Model~2. By comparing the accuracy of Model~1 and Model~2 on the Domain~1 test set, we quantify the influence of Domain~2 training on the model’s previously acquired
reasoning ability in Domain~1. Conversely, by comparing the accuracy of Model~1 and Model~2 on the Domain~2 test set, we measure the extent to which training on Domain~1 affects the model’s subsequent performance in Domain~2. The detailed results of four domains are reported in \Cref{tab:four_domain_one_row}.



\begin{figure*}[t]
  \centering
  \vspace{-0.08in}
  \begin{minipage}[t]{0.48\linewidth}
    \centering
    \includegraphics[width=\linewidth]{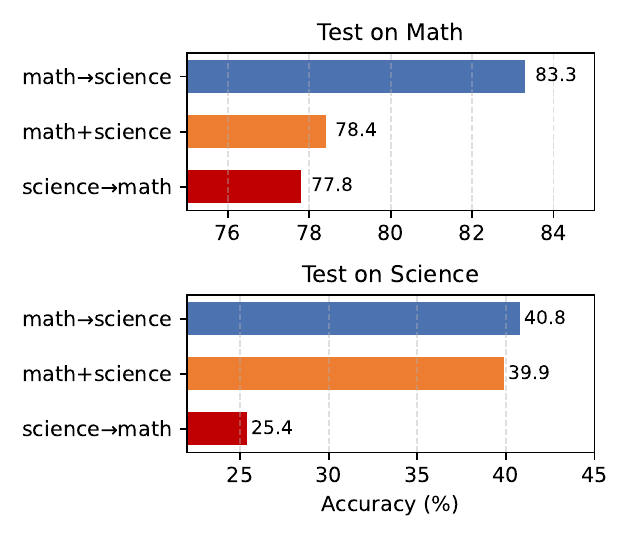}
    \\[-0.4em]
    \label{fig:asymmetric_transfer_math}
  \end{minipage}
  \hfill
  \begin{minipage}[t]{0.48\linewidth}
    \centering
    \includegraphics[width=\linewidth]{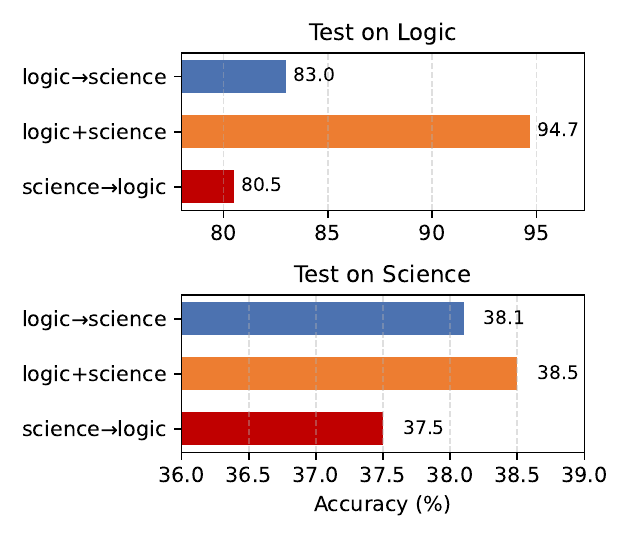}
    \\[-0.4em]
    \label{fig:asymmetric_transfer_logic}
  \end{minipage}
  \vspace{-6pt}
  \caption{
    Comparison of sequential and mixed training between two domains. The left panel compares math and science, while the right panel compares logic and science. Here, math$\rightarrow$science denotes training first on math and then on science, and math+science denotes mixed training on both domains. For math and science, training order is critical, with math$\rightarrow$science achieving the best performance. In contrast, for logic and science, mixed training yields the strongest overall performance.
    }
  \label{fig:asymmetric_transfer}
\end{figure*}
\begin{figure}
  \centering
  \vspace{-0.08in}
  \includegraphics[width=0.9\linewidth]{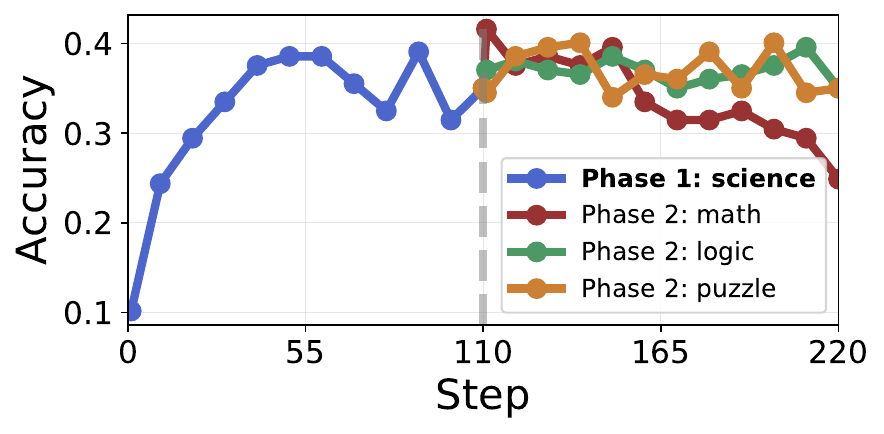}
  \vspace{-6pt}
  \caption{
    Step-wise accuracy on GPQA during two-domain training. In the first stage, the model is trained on science data. In the second stage, training continues with math, logic, or puzzle data. When math data are used in the second stage, GPQA accuracy decreases effectively, indicating that math training interferes with the model’s science reasoning ability.
    }
  \label{fig:science}
\end{figure}

\begin{figure}[t]
  \centering
  \vspace{-0.08in}
  \includegraphics[width=\linewidth]{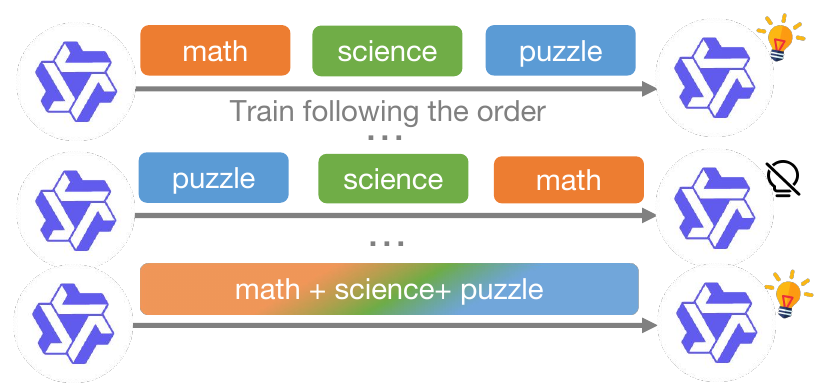}
  \vspace{-6pt}
  \caption{
    The Pipeline of sequential and mixed training over three fields. Top: stage-wise (sequential) training, where the model is trained on math, then science, and finally puzzle. Bottom: mixed training, where data from math/science/logic are combined and used jointly in a single training run.
  }
  \label{fig:seq_vs_mixed}
\end{figure}

\subsection{Results on two-domain training}



\textbf{Results on Math}. Subsequent training on science or logic largely preserves math performance, whereas puzzle training causes noticeable degradation. Conversely, prior science or puzzle training hinders math learning, while prior logic training has little effect.

\textbf{Results on Science}. The science domain exhibits a stronger sensitivity than math. When the model is first trained on science and then trained on math, its science performance is significantly degraded. In contrast, when math training precedes science training, it facilitates scientific reasoning.



\textbf{Results on Logic}. The logic domain exhibits a strong interaction with the science domain. Regardless of whether science training is performed after logic training or before it, science training consistently interferes with the model’s logic reasoning ability, leading to a 5.1\% drop when science is trained afterward and a 7.6\% drop when science is trained beforehand.
In contrast, training on other domains has a relatively minor impact on logic
reasoning performance.


\textbf{Results on Puzzle}. In contrast, the puzzle domain is largely insensitive to cross-domain training, with other domains exerting minimal influence and resulting in neither substantial performance degradation nor noticeable improvement.

\begin{table*}[t]
\caption{
    Performance comparison of sequential and mixed training strategies on math, science and puzzle.  We report final accuracy (\%) on three test domains, together with the corresponding rank (1 = best). Avg denotes the mean accuracy across Math, Science, and Puzzle. Sequential training is denoted as A$\rightarrow$B$\rightarrow$C,
    indicating that the model is trained on A, then B, and finally C. Single-domain refers to models trained only on the corresponding domain (e.g., Math-only, Science-only, or Puzzle-only), evaluated on the same domain.
}
\label{tab:seq_vs_mixed}
\vspace{-6pt}
\centering
\setlength{\tabcolsep}{7pt}
\begin{tabular}{c|cc|cc|cc|cc}
\toprule
\multirow{2}{*}{Training Strategy}
& \multicolumn{2}{c|}{Math}
& \multicolumn{2}{c|}{Science}
& \multicolumn{2}{c|}{Puzzle}
& \multicolumn{2}{c}{Avg} \\
\cmidrule(lr){2-3} \cmidrule(lr){4-5} \cmidrule(lr){6-7} \cmidrule(lr){8-9}
& Acc & Rank & Acc & Rank & Acc & Rank & Acc & Rank \\
\midrule
single-domain
& 82.76 & -- & 38.28 & -- & 20.15 & -- & 47.06 & -- \\

\midrule

math $\rightarrow$ science $\rightarrow$ puzzle
& 80.48 & 3 & 36.97 & 4 & 17.06 & 7 & 44.84 & 4 \\

math $\rightarrow$ puzzle $\rightarrow$ science
& 81.76 & 2 & 39.39 & 2 & 18.94 & 5 & 46.70 & 3 \\

science $\rightarrow$ math $\rightarrow$ puzzle
& 77.68 & 7 & 26.26 & 7 & 20.20 & 2 & 41.38 & 7 \\

science $\rightarrow$ puzzle $\rightarrow$ math
& 78.04 & 6 & 30.20 & 6 & 20.10 & 3 & 42.78 & 6 \\

puzzle $\rightarrow$ math $\rightarrow$ science
& \textbf{82.80} & 1 & 39.29 & 3 & 18.99 & 4 & 47.03 & 2 \\

puzzle $\rightarrow$ science $\rightarrow$ math
& 79.72 & 5 & 32.42 & 5 & 17.72 & 6 & 43.29 & 5 \\

\midrule
mixed (math + science + puzzle)
& 79.92 & 4 & \textbf{41.62} & 1 & \textbf{21.42} & 1 & \textbf{47.65} & 1 \\

\bottomrule
\end{tabular}
\end{table*}
\begin{table*}[t]
\caption{
    Performance comparison of sequential and mixed training strategies on math, science and logic.
}
\label{tab:seq_vs_mixed_logic}
\vspace{-6pt}
\centering
\setlength{\tabcolsep}{7pt}
\begin{tabular}{c|cc|cc|cc|cc}
\toprule
\multirow{2}{*}{Training Strategy}
& \multicolumn{2}{c|}{Math}
& \multicolumn{2}{c|}{Science}
& \multicolumn{2}{c|}{Logic}
& \multicolumn{2}{c}{Avg} \\
\cmidrule(lr){2-3} \cmidrule(lr){4-5} \cmidrule(lr){6-7} \cmidrule(lr){8-9}
& Acc & Rank & Acc & Rank & Acc & Rank & Acc & Rank \\
\midrule
single-domain
& 82.76 & -- & 38.28 & -- & 88.14 & -- & 69.73 & -- \\
\midrule
logic $\rightarrow$ math $\rightarrow$ science
& \textbf{84.00} & 1 & 36.67 & 3 & 87.31 & 2 & 69.99 & 2 \\

logic $\rightarrow$ science $\rightarrow$ math
& 76.96 & 7 & 29.29 & 6 & 81.71 & 4 & 62.65 & 6 \\

math $\rightarrow$ logic $\rightarrow$ science
& 82.56 & 3 & 35.86 & 4 & 83.00 & 3 & 67.14 & 3 \\

math $\rightarrow$ science $\rightarrow$ logic
& 82.80 & 2 & 37.98 & 2 & 75.91 & 5 & 65.56 & 4 \\

science $\rightarrow$ logic $\rightarrow$ math
& 77.96 & 6 & 30.40 & 5 & 76.20 & 6 & 61.52 & 5 \\

science $\rightarrow$ math $\rightarrow$ logic
& 78.24 & 5 & 26.97 & 7 & 63.80 & 7 & 56.34 & 7 \\

\midrule
mixed (math + science + logic)
& 80.44 & 4 & \textbf{40.00} & 1 & \textbf{95.37} & 1 & \textbf{71.94} & 1 \\

\bottomrule
\end{tabular}
\end{table*}

\subsection{Training Order Matters for Math and Science}


We observe that the math and science domains exhibit particularly strong bidirectional interactions and are highly sensitive to the training order. In the following, we focus on a detailed analysis of the relationship between them.


We first examine the Weights \& Biases curves at the value step after science training, followed by continued training on other domains, as shown in \Cref{fig:science}. During training, continued training on the logic and puzzle domains largely
preserves the model’s scientific reasoning ability, whereas training on the math domain leads to a gradual degradation.


In addition, we experiment with jointly training on math and science data. As illustrated in \Cref{fig:asymmetric_transfer}, training on math before science is relatively beneficial for both domains. In contrast, training on science before math simultaneously degrades performance on both math and science. Notably, mixed training on math and science effectively mitigates the negative effects induced by training order, leading to more stable performance across
both domains.

\subsection{Science Interferes with Logical Reasoning}


As shown in \Cref{tab:four_domain_one_row}, in the logic domain, model performance consistently degrades after training on science, regardless of whether science training is performed before or after logic training. After sequential training on logic and science, the model exhibits a noticeable drop in logic accuracy, suggesting that these two domains may be mutually inhibitive: training on science data can damage the logic reasoning ability acquired from logic training, leading to ability forgetting in the logic domain.


We further ask whether mixing logic and science data can mitigate this cross-domain interference. As illustrated in \Cref{fig:asymmetric_transfer}, we compare three training strategies (science$\rightarrow$logic, mixed science+logic, and logic$\rightarrow$science) and evaluate their performance on both logic and science test sets. The science$\rightarrow$logic strategy achieves the lowest accuracy on both domains, while mixed training yields the best performance on both logic and science. These results indicate that jointly training on logic and science can effectively alleviate cross-domain interference and reduce ability forgetting in both of logic and science.

\section{Optimizing Multi-Domain Training Performance}



In this section, we explore how different multi-domain training strategies impact reinforcement learning for reasoning. To this end, we consider two representative domain settings: (1) math, science and puzzle, and (2) math, science, and logic. For the three-domain setting, we examine two training paradigms: sequential training and mixed training. Sequential training admits six possible domain orders, and together with mixed training, results in seven distinct training configurations for comparison.

As illustrated in the pipeline shown in \Cref{fig:seq_vs_mixed}, we first train models using all possible sequential orders of the three domains, followed by a model trained on mixed-domain data. We then evaluate all trained models on the three target domains and compare their accuracies to assess the relative strengths and weaknesses of different training strategies.

\subsection{Training Order Matters in Sequential Training}


As shown in \Cref{tab:seq_vs_mixed} and \Cref{tab:seq_vs_mixed_logic}, when focusing on sequential training, the performance differences across training orders are substantial. For example, the sequence logic$\rightarrow$math$\rightarrow$science achieves the best overall performance, reaching 84\% on math, 36\% on science, and 87\% on logic, with an average accuracy of approximately 70\%. In contrast, the sequence science$\rightarrow$math$\rightarrow$logic leads to severe degradation across all three domains, resulting in the lowest average accuracy of about 56\%.


Across all sequential training strategies, we observe that models containing the
math$\rightarrow$science ordering consistently outperform those following the
reverse science$\rightarrow$math order.
This pattern holds for both the math–science–puzzle and the math–science–logic
settings, indicating that the math-to-science progression constitutes a
critical and favorable training order in multi-domain GRPO.

\subsection{Mixed Training Is Preferable for Certain Domains}



We further compare mixed training with the best-performing sequential training strategy and observe clear domain-dependent preferences. For the math domain, sequential training is more effective and can even outperform training on math alone, suggesting that carefully ordered exposure to auxiliary domains can strengthen mathematical reasoning. In contrast, for science, logic, and puzzle, mixed training consistently yields better results, enabling performance gains beyond those achieved by single-domain training. These findings indicate that different domains benefit from fundamentally different training paradigms, and that a uniform multi-domain strategy may be suboptimal.

\section{related works}

\textbf{LLM Reasoning Ability}. With the rapid development of large language models, reasoning ability
\cite{zhang2023multimodal,yao2023tree,plaat2024reasoning} has emerged
as a core capability and a widely recognized prerequisite for achieving
Artificial General Intelligence (AGI)
\cite{minaee2024large,xu2024survey,feng2024far,krishnan2025artificial}.
Early evidence of strong long-chain reasoning can be traced back to OpenAI o1
\cite{jaech2024openai,arrieta2025o3,hurst2024gpt}, which demonstrated superior
performance on mathematical reasoning benchmarks.
Subsequent models such as QwQ
\cite{qwen2.5,bai2023qwen,bai2023qwens,chu2024qwen2} further advanced reasoning
via process reward modeling
\cite{li2024process,ma2023let,zhang2025lessons,lambert2024rewardbench}.
More recently, DeepSeek R1
\cite{deepseekai2025deepseekr1incentivizingreasoningcapability} and Kimi 1.5
\cite{kimiteam2025kimik15scalingreinforcement} have significantly strengthened
open-source reasoning models, with DeepSeek R1 leveraging simple rule-based
reward mechanisms
\cite{ramesh2024group,hu2025reinforce++,shao2024deepseekmath,alonso2025mathematics}
to close the gap with closed-source systems, and Kimi 1.5 employing techniques
such as long-to-short reasoning for efficiency~\cite{kirk2023understanding,yang2024bayesian}.
In parallel, a line of work including Sky-Thought T1 \cite{sky_t1_2025},
Bespoke-Stratos \cite{bespoke_stratos}, s1
\cite{muennighoff2025s1simpletesttimescaling}, and LIMO
\cite{ye2025limoreasoning} has highlighted the critical role of high-quality,
carefully constructed reasoning data, demonstrating that strong reasoning
performance can be achieved with relatively small but well-designed training
sets.


\textbf{RLVR and GRPO.} With the emergence of DeepSeek-R1, Group Relative Policy Optimization (GRPO)\cite{guo2025deepseek,liu2024deepseek} has
become a widely adopted approach for enhancing reasoning abilities in language models. A growing body of work has focused on improving GRPO itself, including variants such as DAPO~\cite{yu2025dapo}, Dr.\ GRPO~\cite{liu2025understanding}, and GSPO~\cite{zheng2025group}. Other studies investigate entropy collapse in GRPO-based training. For example, NGRPO~\cite{nan2025ngrpo} analyzes how erroneous examples affect entropy during training, while \emph{Rethinking Entropy Interventions in RLVR}~\cite{hao2025rethinking} studies the impact of different token-level behaviors on entropy dynamics from a reinforcement learning perspective. 
In parallel, several works aim to reduce the training cost of GRPO~\cite{xu2025thinking, xu2025scalable}. \emph{It Takes Two}~\cite{wu2025takes} generates paired examples and interprets GRPO through a contrastive learning lens.


\textbf{GRPO for Other Domains.} Beyond mathematics, code is one of the most common domains where GRPO has been applied, as exemplified by models such as Code-R1~\cite{liu2025code}, DeepCoder~\cite{luo2025deepcoder}, and SkyWork OR1~\cite{skywork-or1-2025}. Many of these works jointly train models to acquire both mathematical and coding reasoning abilities. In addition, a growing body of research has explored the application of GRPO to other reasoning domains. For example, Enigmata~\cite{chen2025enigmata} focuses on improving puzzle-solving ability, while Logic-RL~\cite{xie2025logicrlunleashingllmreasoning} targets logical reasoning. More recently, GRPO-style training has also been used to enhance models’ tool-use and agentic capabilities across diverse tasks~\cite{wei2025swe}.

\section{Conclusion}

We show that GRPO exhibits strong, asymmetric, and order-sensitive cross-domain effects in multi-domain reasoning training. Our results demonstrate that different domains favor different training strategies, and that no single paradigm is universally optimal across reasoning domains. In particular, both the choice of domains and their training order can substantially alter learning dynamics and final performance. More broadly, this work highlights that understanding and explicitly managing cross-domain interactions is critical for building robust and generalizable reinforcement learning–based reasoning models, especially as reinforcement learning for reasoning increasingly shifts from single-domain optimization to multi-domain training regimes. We hope our work provides a foundation and practical guidance.


\newpage



\bibliography{example_paper}
\bibliographystyle{icml2026}

\newpage
\appendix
\onecolumn
\section{Limitations}


While this work identifies pronounced asymmetric and order-sensitive cross-domain effects in reinforcement learning for reasoning, it does not provide a deep mechanistic explanation for why such effects arise. In particular, we do not explicitly disentangle whether these behaviors stem from differences in pretraining data coverage, domain-specific knowledge overlap, or optimization dynamics during GRPO training. Understanding the precise origins of cross-domain facilitation and interference—especially their connection to pretraining distributions and representation sharing—remains an important direction for future work.

\section{Impact Statement}

This paper studies the behavior of Group Relative Policy Optimization (GRPO)
under multi-domain reasoning settings, with the goal of improving the
understanding and design of reinforcement learning–based training strategies
for large language models. The primary contribution of this work is analytical
and methodological, focusing on training dynamics, domain interactions, and
performance trade-offs, rather than on deploying models in real-world
applications.

As such, we do not foresee immediate negative societal or ethical consequences
arising directly from this work. On the contrary, by highlighting the
importance of domain-aware and order-aware training, our findings may help
future research develop more robust, interpretable, and reliable reasoning
models. Any broader societal impact of this work will depend on downstream
applications of large language models, which are beyond the scope of this
study.

\section{Test Performance on Other Two-domain Training}

\begin{figure*}[ht]
  \centering
  \vspace{-0.1in}
  \begin{minipage}[t]{0.32\textwidth}
    \centering
    \includegraphics[width=\linewidth]{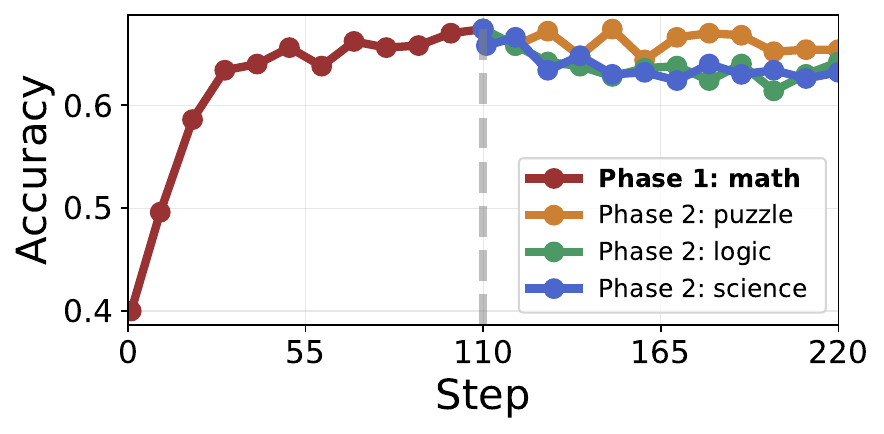}
    \\[-0.3em]
    {(a) MATH500}
  \end{minipage}
  \hfill
  \begin{minipage}[t]{0.32\textwidth}
    \centering
    \includegraphics[width=\linewidth]{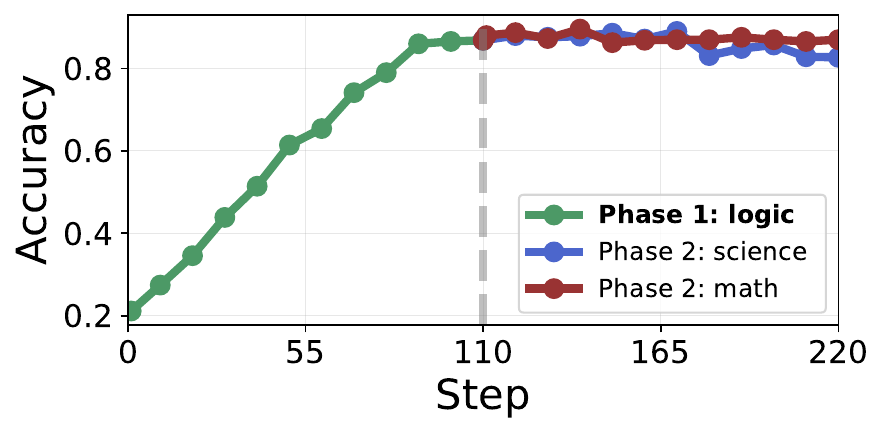}
    \\[-0.3em]
    { (b) logic}
  \end{minipage}
  \hfill
  \begin{minipage}[t]{0.32\textwidth}
    \centering
    \includegraphics[width=\linewidth]{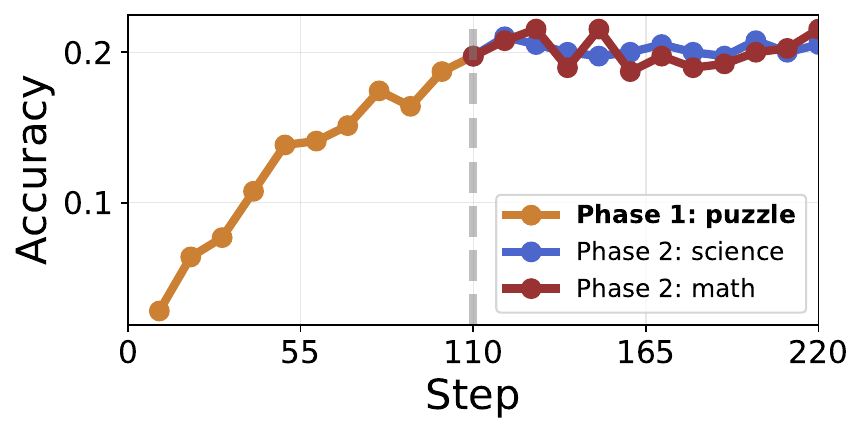}
    \\[-0.3em]
    {(c) puzzle}
  \end{minipage}
  \vspace{-0.05in}
  \caption{
    \textbf{Two-domain training results on science, logic, and puzzle.}
    Accuracy curves across training steps under different two-domain training
    strategies.
    }
  \label{fig:mix_others}
  \vspace{-6pt}
\end{figure*}


Here we show the step-wise accuracy of the Qwen3-4B-Base model under two-domain
training across three domains, with all curves directly obtained from
Weights \& Biases logs.

\section{Motivation: Single-domain v.s. Multi-domain GRPO}

Given a prompt $x$, Group Relative Policy Optimization (GRPO) samples a group of
$K$ responses $\{y_i\}_{i=1}^K \sim \pi_{\theta}(\cdot\mid x)$ and assigns scalar
rewards $r_i = R(x,y_i)$.
A group-relative advantage is computed by normalizing rewards within the group:
\begin{equation}
A_i = \frac{r_i - \mu}{\sigma + \epsilon},
\qquad
\mu = \frac{1}{K}\sum_{j=1}^{K} r_j,
\end{equation}
where $\sigma$ denotes the standard deviation of rewards within the group.
The policy is updated by optimizing a PPO-style clipped objective. In its standard formulation, GRPO implicitly assumes a \emph{single-domain}
setting, where rewards are generated from a homogeneous task distribution.
Under this assumption, group-relative normalization is applied over responses
drawn from the same reasoning domain, and the resulting policy updates optimize a
single domain-specific reasoning objective.


\paragraph{Multi-domain GRPO.} In practice, reinforcement learning for reasoning is often performed over data from multiple domains, such as mathematics, science, logic, and puzzles.
Let $\mathcal{D}$ denote the set of domains and $p(d)$ a domain mixing
distribution.
Training under this setting optimizes the objective
\begin{equation}
\mathcal{L}(\theta)
=
\sum_{d\in\mathcal{D}} p(d)\,
\mathbb{E}_{x\sim\mathcal{X}_d}
\big[\mathcal{L}_{\mathrm{GRPO}}(\theta; x)\big],
\end{equation}
where $\mathcal{L}_{\mathrm{GRPO}}(\theta; x)$ follows the single-domain GRPO
objective. Unlike the single-domain case, rewards in the multi-domain setting originate from heterogeneous domain distributions and may encode different reasoning objectives. As a result, policy updates aggregate domain-specific advantage signals within
the same optimization process. 

This raises a fundamental question: \textit{do rewards from different reasoning domains mutually facilitate the acquisition of reasoning abilities, or do they interfere with previously learned capabilities during reinforcement learning?} In this work, we empirically investigate this question by analyzing cross-domain transfer and interference under multi-domain.


\section{Training Details of Experiments}


\lstdefinestyle{yamlstyle}{
  frame=single,
  rulecolor=\color{black!40},
  basicstyle=\ttfamily\scriptsize,
  lineskip=0.5pt,
  xleftmargin=4pt,
  xrightmargin=4pt,
  showstringspaces=false,
  breaklines=true,
  breakatwhitespace=true
}

\vspace{-4pt}
\begin{lstlisting}[style=yamlstyle]
# Key training configuration (GRPO / RLVR)
run:
  project: MIX-REWARD-GRPO-SMALL-Normal-Long
  experiment: vanilla-Qwen3-4B-Base-msl
  seed: 0
  total_epochs: 6

model:
  base: Qwen/Qwen3-4B-Base
  max_prompt_len: 1024
  max_response_len: 16384
  remove_padding: true
  grad_checkpointing: true

algorithm:
  name: GRPO
  loss_mode: vanilla
  loss_agg: token-mean
  n_resp_per_prompt: 8
  clip_ratio_low: 3.0e-4
  clip_ratio_high: 4.0e-4
  entropy_coeff: 0.0
  grad_clip: 1.0
  kl_in_reward: false
  kl_coef: 0.0
  kl_loss: false
  kl_loss_coef: 0.0

optimization:
  lr: 1.0e-6
  lr_warmup_ratio: 0.05
  weight_decay: 0.1
  train_batch_size: 256
  mini_batch_size: 64
  micro_batch_size_per_gpu: 16
  dynamic_bsz: true

rollout:
  engine: vllm
  mode: sync
  gpu_mem_util: 0.6
  tensor_parallel: 1
  chunked_prefill: true
  max_num_batched_tokens: 17408

compute:
  nnodes: 1
  gpus_per_node: 8
  ray_num_cpus: 64
  sequence_parallel: 1
  param_offload: false
  optimizer_offload: false
  entropy_checkpointing: true
\end{lstlisting}
\vspace{-6pt}

\end{document}